\documentclass[runningheads]{llncs}
\usepackage{graphicx}
\usepackage{times}
\usepackage{epsfig}
\usepackage{times}
\usepackage{amsmath}
\usepackage{hyperref}
\usepackage{amssymb}
\usepackage{tikz} % To generate the plot from csv
\usepackage{xcolor}
\usepackage{pgfplots}
\usepackage{caption}

\usepackage{xcolor}

\newcommand{\etal}{\textit{et al}.}
\newcommand{\ie}{\textit{i}.\textit{e}. }
\newcommand{\eg}{\textit{e}.\textit{g}. }

\begin{document}

\title{Performance Evaluation of Learned 3D Features}
\author{Riccardo Spezialetti \and
Samuele Salti \and
Luigi Di Stefano }
\authorrunning{R.Spezialetti et al.}
\institute{Viale del Risorgimento 2, Bologna, Italy
\email{\{riccardo.spezialetti,samuele.salti,luigi.distefano\}@unibo.it}}
\maketitle              % typeset the header of the contribution
\begin{abstract}
	Matching surfaces is a challenging 3D Computer Vision problem typically addressed by local features. Although a variety of 3D feature detectors and descriptors has been proposed in literature, they have seldom been proposed together and it is yet not clear how to identify the  most effective detector-descriptor pair for a specific application. A promising solution is to leverage machine learning to learn the optimal 3D detector for any given 3D descriptor \cite{tonioni2018learning}. In this paper, we report a performance evaluation of the detector-descriptor pairs obtained by learning a paired 3D detector for the most popular 3D descriptors. In particular, we address experimental settings dealing with object recognition and surface registration.	
\end{abstract}

\section{Introduction}
\label{intro}
Surface matching is an ubiquitous task in 3D Computer Vision, where it helps to tackle major applications such as object recognition and surface registration.
Nowadays, most surface matching methods follow a \textit{local} paradigm based on establishing correspondences between 3D patches referred to as \textit{features}. The typical \textit{feature-matching pipeline} consists of three steps: \textit{detection}, \textit{description} and \textit{matching}.

Although over the last decades many 3D detectors and descriptors have been proposed in literature, it is yet unclear how to effectively combine these proposals to create an effective pipeline. Indeed, unlike the related field of \textit{local image features}, methods to either detect or describe 3D features have been designed and proposed separately, alongside with specific application settings and related datasets. This is also vouched by the main performance evaluation papers in the field, which address either repeatability of 3D detectors designed to highlight geometrically salient surface patches \cite{tombari2013performance} or distinctiveness and robustness of  popular 3D descriptors \cite{guo2016comprehensive}. 

More recently, however, \cite{salti2015learning} and \cite{tonioni2018learning} have proposed 
a machine learning approach that allows for learning an optimal 3D keypoint detector for any given 3D descriptor so as to maximize the end-to-end performance of the overall \textit{feature-matching pipeline}.  
The authors show that this approach provides effective pipelines across diverse applications and datasets. Moreover, their object recognition experiments show that, with the considered descriptors   (\textit{SHOT} \cite{Tombari:2010:USH:1927006.1927035}, \textit{Spin Image (SI)}  \cite{johnson1999using}, \textit{FPFH}  \cite{rusu2010fast}), learning to detect specific keypoints leads to better performance than relying on existing general-purpose handcrafted detectors (\textit{ISS} \cite{zhong2009intrinsic}, \textit{Harris3D} \cite{sipiran2011harris}, \textit{NARF} \cite{steder2010narf}).   

By enabling an optimal detector to be learned for any descriptor, \cite{tonioni2018learning} sets forth a novel paradigm to maximize affinity between 3D detectors and descriptors. This opens up the question of which learned detector-descriptor pair may turn out most effective in the main application areas. This paper tries to answer this question by proposing an experimental evaluation of learned 3D pipelines. In particular, we address object recognition and surface registration, and compare the performance attained by learning a paired feature detector for the most popular handcrafted 3D descriptors (\textit{SHOT} \cite{Tombari:2010:USH:1927006.1927035}, \textit{SI}  \cite{johnson1999using}, \textit{FPFH}  \cite{rusu2010fast}, \textit{USC}\cite{tombari2010unique}, \textit{RoPS} \cite{guo2013rotational}) as well as for a recently proposed descriptor based on deep learning (\textit{CGF-32} \cite{khoury2017learning}).  

\section{3D Local Feature Detectors and Descriptors}
\label{Review3DFeatures}
This section reviews state-of-the-art methods for detection and description of 3D local features. Both tasks have been pursued through \textit{hand-crafted} and \textit{learned} approaches.

\paragraph{Hand-Crafted Feature Detectors}
Keypoint detectors have traditionally been conceived to identify points that maximize a saliency function computed on a surrounding patch. The purpose of this function is to highlight those local geometries that turn out repeatedly identifiable in presence of nuisances such as noise, viewpoint changes, point density variations and clutter. State-of-the-art proposals mainly differ for the adopted saliency function. Detectors operate in two steps: first, the saliency function is computed at each point on the surface, then \textit{non-maxima suppression} allows for sifting out saliency peaks. 
\textit{Intrinsic Shape Signature (ISS)} \cite{zhong2009intrinsic} computes the eigenvalue decomposition of the scatter-matrix of the points within the supporting patch in order to highlight local geometries exhibiting a prominent principal direction, \textit{Harris3D} \cite{sipiran2011harris} extends the idea of image corners  by deploying surface normals rather than image gradients to calculate the saliency (i.e \textit{Cornerness}) function. 
\textit{Normal Aligned Radial Feature (NARF)} \cite{steder2010narf} first selects stable surface points, then highlights those stable points showing sufficient local variations. This leads to locate keypoints close to depth discontinuities.

\paragraph{Learned Feature Detectors}
Unlike previous work in the field, Salti \etal{} \cite{salti2015learning} proposed to learn a keypoint detector amenable to identify points likely to generate correct matches when encoded by the \textit{SHOT} descriptor. In particular, the authors cast keypoint detection as a binary classification problem tackled by a Random Forest and show how to generate the training set as well as the feature representation deployed by the classifier. Later, Tonioni \etal{} \cite{tonioni2018learning} have demonstrated that this approach can be applied seamlessly and very effectively to other popular descriptors such as   \textit{SI}  \cite{johnson1999using} and  \textit{FPFH} \cite{rusu2010fast}.

\paragraph{Hand-Crafted Feature Descriptors}
Many hand-crafted feature descriptors represent the local surface by computing geometric measurements within the supporting patch and then accumulating values into histograms. \textit{Spin Images (SI)} \cite{johnson1999using} relies on two coordinates to represent each point in the support: the radial coordinate, defined as the perpendicular distance to the line trough the surface normal at the keypoint, and the elevation coordinate, defined as the signed distance to the tangent plane at the keypoint. The space formed by this two values is then discretized into a 2D histogram.

In \textit{3D Shape Context (3DSC)} \cite{frome2004recognizing} the support is partitioned by a 3D spherical grid centered at the keypoint with the north pole aligned to the surface normal. A 3D histogram is built by counting up the weighted number of points falling into each spatial bin along the radial, azimuth and elevation dimensions.
\textit{Unique Shape Context (USC)}\cite{tombari2010unique} extends \textit{3DSC} with the introduction of a unique and repeatable canonical reference frame borrowed from \cite{Tombari:2010:USH:1927006.1927035}.

\textit{SHOT} \cite{Tombari:2010:USH:1927006.1927035}, alike, deploys both a unique and repeatable canonical reference frame as well as a 3D spherical grid to discretize the supporting patch into bins along the radial, azimuth and elevation axes. Then, the angles between the normal at the keypoint and those at the neighboring points within each bins are accumulated into local histograms.
\textit{Rotational Projection Statistics (RoPS)} \cite{guo2013rotational} uses a canonical reference frame to rotate the neighboring points on the local surface. The descriptor is then constructed by rotationally projecting the 3D points onto 2D planes to generate three distribution matrices. Finally, a  histogram encoding five statistics of distribution matrices is calculated. \textit{Fast Point Feature Histograms (FPFH)} \cite{rusu2010fast} operates in two steps. In the first, akin to PFH \cite{rusu2008aligning}, four features, refereed to as SPFH, are calculated using the Darboux frame and the surface normals between the keypoint and its neighbors. In the second step, the descriptor is obtained as the weighted sum between the SPFH of the keypoint and the SPFHs of the neighboring points. 

\paragraph{Learned Feature Descriptors}
The success of deep neural networks in so many challenging image recognition tasks has motivated research on learning representations from 3D data.
One of the pioneering works 
is \textit{3D Match} \cite{zeng20173dmatch}, where the authors deploy a siamese network trained on local volumetric patches to learn  a local descriptor. The input to the network consists of a Truncated Signed Distance Function (TSDF) defined on a voxel grid. In \cite{khoury2017learning}, the authors deploy a fully-connected deep neural network together with a feature learning approach based on the \textit{triplet ranking loss}
in order to learn a very compact 3D descriptor, referred to as \textit{CGF-32}. Their approach  does not rely on raw data but on an hand-crafted input representation similar to \cite{frome2004recognizing}, canonicalized by the local reference frame presented in \cite{Tombari:2010:USH:1927006.1927035}.

\section{Keypoint Learning}
\label{keypointlearning}
In order to carry out the performance evaluation proposed in this paper, for most local descriptors reviewed in \autoref{Review3DFeatures} we did learn the corresponding optimal detector according to the \textit{keypoint learning} methodology \cite{tonioni2018learning}. We provide here a brief overview of this methodology and  refer the reader to \cite{salti2015learning,tonioni2018learning} for a detailed description.

The idea behind keypoint learning is to learn to detect keypoints that can yield good correspondences when coupled with a given descriptor. To this end, keypoint detection is cast as binary classification, \textit{i.e.} a point can either be a good candidate or not when used to create matches by means of the given descriptor, and a Random Forest is used as classifier. Training of the classifier requires to define the training set, \textit{i.e.} both positive (good) and negative (not good) points, as well as the feature representation.

As for positive samples, the method tries to sift out those points that,  when described by a chosen descriptor, can be matched correctly across different 2.5D views of a 3D object. Thus, starting from a set of  2.5D views $\{V_i\}, i=1,\dots,N$ of an object from a 3D dataset, each point $p \in V_i$ in each view $V_i$ is embedded by the chosen descriptor. Then, for each view $V_i$, a subset of overlapping views is selected based on an overlap threshold $\tau$. A two-step positive samples selection is performed on  $V_i$ and each overlapping view $V_j$. In the first step, a list of correspondences between descriptors is created by searching for all descriptors $d \in V_i$ the nearest neighbor in the descriptor space between all descriptors $g \in V_j$.
A preliminary list of positive samples $P_i^j$ for view $V_i$ is created by taking only those points that have been correctly matched in $V_j$, \ie{} the points belonging to the matched descriptors in the two views correspond to the same 3D point of the object according to threshold $\epsilon$. 
The list is then filtered removing \textit{non-maxima} local extrema within $\epsilon_{nms}$ using the descriptor distance as saliency.
In the second step, the list of positive samples is refined by keeping only the points in $V_i$ that can be  matched correctly also in those others overlapping views that have not been used in the first step.
Negative samples are then extracted on each view, sampling random points among those points which are not included in the positive set.
A distance threshold $\epsilon_{neg}$ is used to avoid a negative being too close to a positive and to other negative samples, and also to balance the size of the positive and negative sets.

As far as the representation input to the classifier is concerned, the method relies on histograms of normal orientations inspired by \textit{SHOT} \cite{Tombari:2010:USH:1927006.1927035}. However, to avoid computation of the local Reference Frame while still achieving rotation invariance, the spherical support is divided only along the radial dimension so as to compute a histogram for each spherical shell thus obtained. \cite{tonioni2018learning} showed that, although inspired by SHOT, such representation can be used to learn an effective detector also for other descriptors.

\section{Evaluation Methodology}
The performance evaluation proposed in this paper aims to compare different learned detector-descriptor pairs while addressing two main application settings, namely object recognition and surface registration. In this section, we  highlight the key traits and nuisances which characterize the two tasks, present the datasets and performance evaluation metrics used in the experiments and, finally, provide the relevant implementation details.  

\subsection{Object Recognition}
In typical object recognition settings, one wishes to recognize a set of given 3D models into scenes acquired from an unknown vantage point and featuring an unknown arrangement of such models. Peculiar nuisances in this scenario are occlusions and, possibly, clutter, as objects not belonging to the model gallery may be present in the scenes. 
In our experiments we rely on the following popular object recognition datasets:
 
\begin{itemize}
\item \textit{UWA} dataset, introduced by Mian \etal  \cite{mian2010repeatability}. This dataset consists of 4 full 3D models and 50 scenes wherein models significantly occlude each other. To create some clutter, scenes contain also an object which is not included in the model gallery. As scenes are scanned by a Minolta Vivid 910 scanner, they are corrupted by real sensor noise. 

\item \textit{Random Views} dataset, based on the Stanford 3D scanning repository \footnote{3 http://graphics.stanford.edu/data/3Dscanrep/} and originally proposed in \cite{tombari2013performance}. 
This dataset comprises 6 full 3D models and 36 scenes obtained by synthetic renderings of random model arrangements. Scenes feature occlusions but no clutter. Moreover, scenes are corrupted by different levels of synthetic noise. In the experiments we consider scenes with Gaussian noise equal to $\sigma=0.1$ mesh resolution units. 
\end{itemize}

To evaluate the effectiveness of the considered learned detector-descriptor pairs we rely on descriptor matching experiments. Specifically, for both datasets, we run keypoint detection on synthetically rendered views of all models. Then, we compute and store into a single kd-tree all the corresponding descriptors. Keypoints are detected and described also in the set of scenes provided with the dataset, $\{S_j\}, j=1,\dots,N_S$. Eventually, a correspondence is established for each scene descriptor by finding the nearest neighbor descriptor within the models kd-tree and thresholding the distance between descriptors to accept a match as valid. Correct correspondences can be identified based on knowledge of the ground-truth transformations which bring views and scenes into a common reference frame and checking whether the matched keypoints lay within a 3D distance $\epsilon$. Indeed, denoting as $(k_j, k_{n,m})$ a correspondence between a keypoint $k_j$ detected in scene $S_j$ and a keypoint $ k_{n,m}$ detected in the $n$-th view of model $m$, as $\textbf{T}_{j,m}$ the transformation from $S_j$ to model $m$, as $\textbf{T}_{n,m}$ the transformation from the $n$-th view and the canonical reference frame of model $m$, the set of correct correspondences for scene $S_j$ is given by: 

\begin{equation}
\mathcal{C}_{j} = \{ (k_j, k_{n,m})  : \| \mathbf{T}_{j,m}k_j - \mathbf{T}_{n,m}k_{n,m}  \| \le \epsilon \}
\end{equation}

From $\mathcal{C}_{j}$, we can compute True Positive and False Positive matches for each scene and, by averaging them across scenes, for each of the considered datasets. The final results for each dataset are provided as \textit{Recall vs. 1-Precision} curves, with curves obtained by varying the threshold on the distance between descriptors. 

\subsection{Surface Registration}
\label{SurfaceRegistrationMethodology}
The goal of surface registration is to align into a common 3D reference frame several partial views (usually referred to as scans) of a 3D object obtained by a certain optical sensor. This is achieved through rather complex procedures that, however, typically rely on a key initial step, referred to as \textit{Pairwise Registration}, aimed at estimating the rigid motion between any two views by a \textit{feature-matching pipeline}. %Thus, in surface registration, 3D feature detection, description and matching are instrumental to attain an as good as possible set of pairwise alignments between the views which then undergoes further processing to get the final global alignment. 
Differently from object recognition scenarios, the main nuisances deal with missing regions, self-occlusions, limited overlap area between views and point density variations. In our experiments we rely on the following surface registration dataset:  

\begin{itemize}
\item \textit{Laser Scan} dataset, recently proposed in \cite{khoury2017learning}. This dataset includes 8  public-domain 3D models, \ie{} 3 taken from the AIM@SHAPE repository (\textit{Bimba}, \textit{Dancing Children} and \textit{Chinese Dragon}), 4 from the Stanford 3D Scanning Repository (\textit{Armadillo}, \textit{Buddha}, \textit{Bunny}, \textit{Stanford Dragon}) and  \textit{Berkeley Angel}  %\cite{kolluri2004spectral}. 
According to the protocol described in \cite{khoury2017learning}, training 
should be carried out based on synthetic views generated from  \textit{Berkeley Angel}, \textit{Bimba}, \textit{Bunny} and \textit{Chinese Dragon}, while the test data consists of the the real scans available for the remaining 3 models (\textit{Armadillo}, \textit{Buddha} and \textit{Stanford Dragon}).
\end{itemize}

Thus, given a set of $M$ real scans available for a test model, we compute all the possible $N = \frac{M (M-1)}{2}$ view pairs $\{V_i, V_j\}$. For each pair, we run keypoint detection on both views. Due to partial overlap  between the views, a keypoint belonging to $V_i$ may have no correspondence in $V_j$. Hence, denoted as  $\textbf{T}_i$ and $\textbf{T}_j$  the ground-truth transformations that, respectively, bring $V_i$ and $V_j$ into a canonical reference frame, we can compute the set $\mathcal{O}_{i,j}$ that contains the keypoints in $V_i$ that have a corresponding point in $V_j$. In particular, given a keypoint $k_i \in V_i$
\begin{equation}
\mathcal{O}_{i,j} = \{ k_i : \| \mathbf{T}_ik_i - \mathcal{NN}(\mathbf{T}_ik_i, \mathbf{T}_jV_j)\| \le \epsilon_{ovr}\} \mbox{,}
\end{equation}

where $\mathcal{NN}(\mathbf{T}_ik_i, \mathbf{T}_jV_j)$ denotes the nearest neighbor of $\mathbf{T}_ik_i$ in 
the transformed view $\textbf{T}_jV_j$.
If the number of points in $\mathcal{O}_{i,j}$ is less than 20\% of the keypoints in $V_i$, the pair $(V_i,V_j)$ is not considered in the evaluation experiments due to insufficient overlap. Conversely, for all the view pairs $(V_i,V_j)$   exhibiting sufficient overlap, a list of correspondences between all the keypoints detected in $V_i$ and all the keypoints extracted from $V_j$ is established by finding the nearest neighbor in the descriptor space via kd-tree matching. Then, given a pair of matched keypoints 
$(k_i,k_j)$, $k_i \in V_i,k_j \in V_j$, the set of correct correspondences, $\mathcal{C}_{i,j}$, can be identified based on the available ground-truth transformations by checking whether the matched keypoints lay within a certain distance $\epsilon$ in the canonical reference frame: 

\begin{equation}
\mathcal{C}_{i,j} = \{ (k_i ,k_j) : \| \mathbf{T}_ik_i - \mathbf{T}_jk_j \| \le \epsilon \}
\end{equation}

Then, the \textit{precision} of the matching process can be computed as a function of the distance threshold $\epsilon$ \cite{khoury2017learning}: 

\begin{equation}
precision_{i,j}(\epsilon) = \frac{ \left |\mathcal{C}_{i,j} \right |} {\left |\mathcal{O}_{i,j} \right |}
\end{equation}

The \textit{precision} score associated with any given model is obtained by averaging across all view pairs. We also average across all test models so as to get the final score associated to the \textit{Laser Scan} dataset.

%Table for Object recognition experiment%
\begin{table*}[ht]
\caption{Parameters for object recognition datasets.}
\label{tab:parameters_obj_rec}
\centering
\resizebox{.90\textwidth}{!}{
\begin{tabular}{l|c|c|c|c|c|c|c|c}
Dataset & $r_{desc}(mm)$ & $r_{det}(mm)$ & $\tau$ & $\epsilon(mm) $ &$\epsilon_{nms}(mm) $ & $\epsilon_{neg}(mm) $ & $r_{nms}(mm)$ & $s_{min}(mm)$\\ \hline
\emph{UWA} 					& 40    & 20    & 0.85 & 7     & 4 & 2 & 4 & 0.8 \\
\emph{Random Views} 		& 40    & 20    & - & 7     & - & - & 4  & 0.8 \\
\end{tabular}}
\end{table*}

%Table for Object recognition experiment%
\begin{table}[ht]
\caption{Parameters for surface registration dataset.}
\label{tab:parameters_surf_reg}
\centering
\resizebox{.90\textwidth}{!}{
\begin{tabular}{l|c|c|c|c|c|c|c|c|c}
Model Name & $r_{desc}(mm)$ & $r_{det}(mm)$ & $\tau$ & $\epsilon(mm) $ &$\epsilon_{nms}(mm) $ & $\epsilon_{neg}(mm) $ & $\epsilon_{ovr} $ & $r_{nms}(mm)$ & $s_{min}(mm)$\\ \hline
\emph{Angel} 			& 40    	& 20   	& 0.85 		& 7     & 4 		& 2		& -		& - 	& - \\
\emph{Bimba} 			& 40    	& 20   	& 0.85 		& 7     & 4 		& 2 	& -		& -	 	& - \\
\emph{Bunny} 			& 40 		& 20 	& 0.65    	& 7    	& 4  		& 2 	& - 	& - 	& -  \\
\emph{Chinese Dragon} 	& 40 		& 20 	& 0.65    	& 7    	& 4  		& 2 	& - 	& - 	& -  \\

\emph{Armadillo}		& 40 		& 20 	& -    		& 7 	& -  		& -  	& 2 	& 4 	& 0.5 \\
\emph{Buddha} 			& 40 		& 20 	& -    		& 7 	& -  		& -  	& 2 	& 4 	& 0.5 \\
\emph{Stanford Dragon} 	& 40 		& 20 	& -    		& 7 	& -  		& -  	& 2 	& 4 	& 0.5 \\
\end{tabular}
}
\end{table}

\subsection{Implementation}
%HW and WS%
For all handcrafted descriptors considered in our evaluation, we use the implementation available in the PCL library. For \textit{CGF-32}, we use the public implementation made available by the authors \cite{khoury2017learning}. As for the \textit{Keypoint Learning} (KPL) framework described in \autoref{keypointlearning}, we use the publicly available original code for the generation of the training set \footnote{http://github.com/CVLAB-Unibo/Keypoint-Learning.}. 
During the detection phase, each point of a point cloud is passed through the Random Forest classifier which produces a score. A point is identified as a keypoint if it exhibits a local maximum of the scores in a neighborhood of radius $r_{nms}$ and the score is higher than a threshold $s_{min}$.
For each descriptor considered in our evaluation, we train its paired detector according to the KPL framework. As a result, we obtain six detector-descriptor pairs, referred to from now on as \textit{KPL-CGF32}, \textit{KPL-FPFH}, \textit{KPL-ROPS}, \textit{KPL-SHOT}, \textit{KPL-SI}, \textit{KPL-USC}.

In object recognition experiments, the training data for all detectors are generated from the 4 full 3D models present in the \textit{UWA} dataset. According to the KPL methodology  \cite{salti2015learning,tonioni2018learning}, for each model we render views from the nodes of an icosahedron centered at the centroid. 

Then, the detectors are used in the scenes of the \textit{UWA} dataset as well as in those of the \textit{Random Views} dataset. Thus, similarly to \cite{salti2015learning,tonioni2018learning}, we do not retrain the detectors on \textit{Random Views} in order to test the ability of the considered detector-descriptor pairs to generalize well to unseen models in object recognition settings. 
A coherent approach was pursued for the \textit{CGF-32} descriptor. As the authors do not provide a model trained on the \textit{UWA} dataset, we trained the descriptor on the synthetically rendered views of the 4  \textit{UWA} models using the code provided by the authors and following the protocol described in the paper in order to generate the data needed by their learning framework based on the \textit{triplet ranking loss}. Thus, \textit{KPL-CGF32} was trained on \textit{UWA} models and, like all other detector-descriptor pairs, tested on both \textit{UWA} and \textit{Random Views} scenes.  

%Registration%
In surface registration experiments we proceed according to the protocol proposed in \cite{khoury2017learning}. Hence, detectors are trained with rendered views of the train models provided within the \textit{Laser Scanner} dataset (\textit{Angel}, \textit{Bimba}, \textit{Bunny}, \textit{Chinese Dragon}) and tested on the real scans of the test models (\textit{Armadillo}, \textit{Buddha}, \textit{Stanford Dragon}).
As \textit{CGF-32} was trained exactly on the abovementioned train models \cite{khoury2017learning}, to carry out surface registration experiments we did not retrain the descriptor but used the trained network published by the authors\footnote{https://github.com/marckhoury/CGF}.

%Test%
The values of the main parameters of the detector-descriptor pairs used in the experiments are summarized in \autoref{tab:parameters_obj_rec} and \autoref{tab:parameters_surf_reg}.
As it can be observed from \autoref{tab:parameters_obj_rec}, train parameters for \textit{Random Views} dataset are not specified as we did not train KPL detectors on this dataset.
For surface registration, since models belong to different repositories, we report parameters grouped by model. Test parameters for \textit{Angel}, \textit{Bimba}, \textit{Bunny} and  \textit{Chinese Dragon} are not reported as they are only used in train. Similarly, we omit train parameters for \textit{Armadillo}, \textit{Buddha} and \textit{Stanford Dragon}. Due to the different shapes of the models in the dataset,  $\tau$ is tuned during the train stage so that the number of overlapping views remains constant across all models. 

\section{Experimental Results}
\subsection{Object Recognition}
\begin{figure}[h]
	\centering
	\begin{tabular}{cc}
		\includegraphics[width=.5\textwidth]{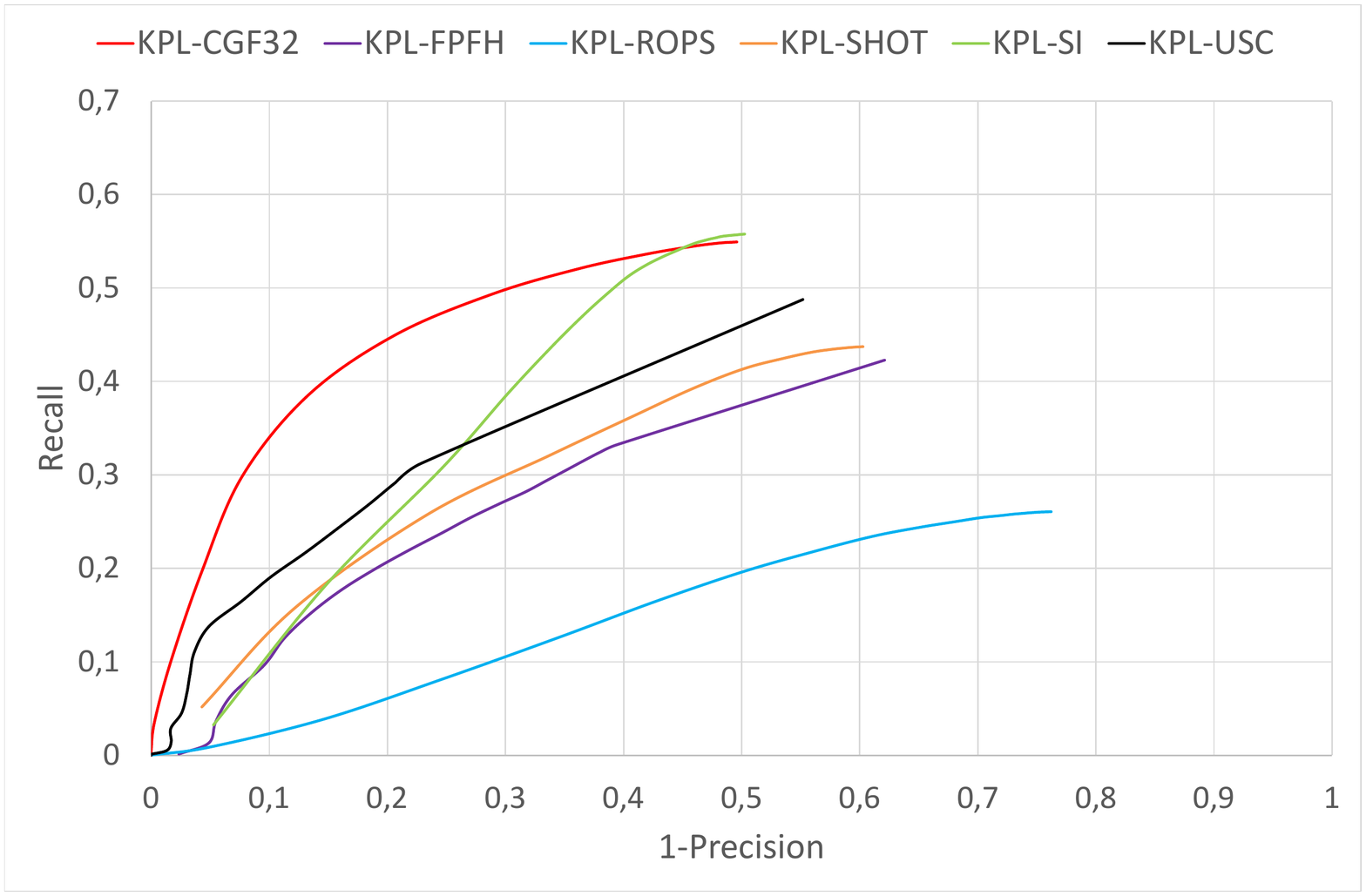} \label{fig:ObjRecResults_a}&
		\includegraphics[width=.5\textwidth]{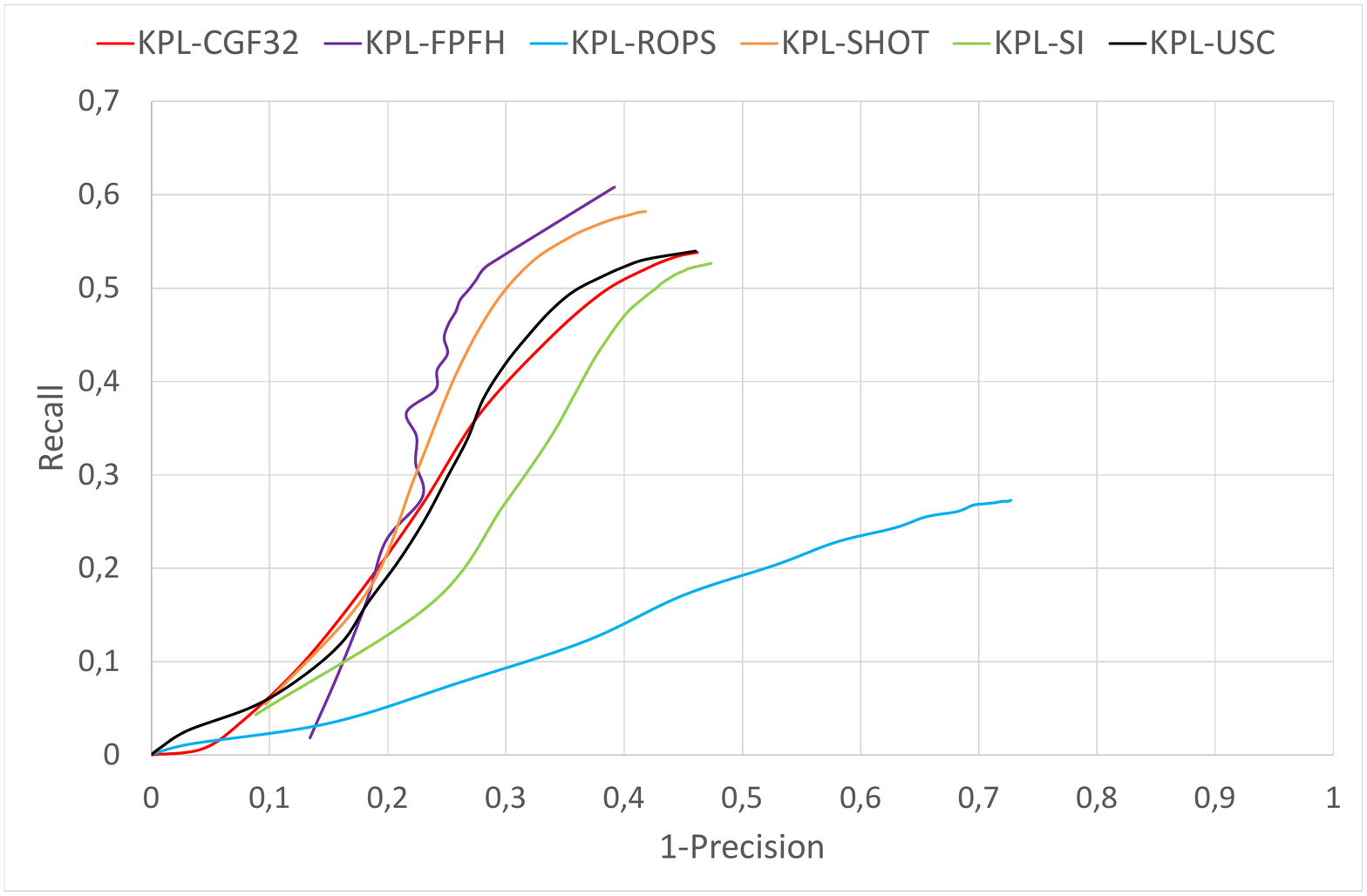} \label{fig:ObjRecResults_b}\\
		(a) &(b) \\
	\end{tabular}
	\caption{Quantitative results on Object recognition. Column a: \textit{UWA} dataset. Column b: \textit{Random Views} dataset.}
	\label{fig:ObjRecResults}
\end{figure}

Results on the \textit{UWA} dataset are shown in \autoref{fig:ObjRecResults}. First, we wish to highlight how the features based on descriptors which encode just the spatial densities of points around a keypoint outperform those relying on higher order geometrical attributes (such as, \eg, normals). Indeed, \textit{KPL-CGF32}, \textit{KPL-USC} and \textit{KPL-SI} yield significantly better results than \textit{KPL-SHOT} and \textit{KPL-FPFH}. These results are coherent with the findings and analysis reported in the performance evaluation by Guo \etal \cite{guo2016comprehensive}, which pointed out the former feature category being  more robust to clutter and sensor noise. It is also worth observing how the representation based on the spatial tessellation and point density measurements proposed in \cite{frome2004recognizing} together with the local reference frame proposed in \cite{Tombari:2010:USH:1927006.1927035} turn out particularly amenable to object recognition, as it is actually deployed by both features yielding neatly the best performance, namely \textit{KPL-CGF32} and \textit{KPL-USC}. Yet, learning a dataset-specific non-linear mapping by a deep neural network on top of this good representation does improve performance quite a lot, as vouched by \textit{KPL-CGF32} outperforming \textit{KPL-USC} by a large margin. Indeed, the results obtained in this paper by learning both a dataset-specific descriptor as well as its paired optional detector, \ie{} the features referred to as \textit{KPL-CGF32}, turn out significantly superior to those previously published on \textit{UWA} object recognition dataset (see \cite{salti2015learning} and \cite{tonioni2018learning}).

In \cite{tonioni2018learning}, the results achieved on \textit{Random Views} by the detectors trained on \textit{UWA} prove the ability of the KPL methodology to learn to detect general rather than dataset-specific local shapes amenable to provide good matches alongside with the paired descriptor, and even more effectively, in fact, than the shapes found by handcrafted detectors. 
Thus, when comparing the different features, we can assume here that descriptors are feed by detectors with optimal patches and focus on the ability of the former to handle the specific nuisances of the \textit{Random Views} dataset. As shown in \autoref{fig:ObjRecResults}, \textit{KPL-FPFH} and \textit{KPL-SHOT} perform slightly better than \textit{KPL-USC}, \textit{KPL-CGF32} and \textit{KPL-SI}. Again, this is coherent with previous findings reported in literature (see  \cite{guo2016comprehensive} and \cite{tonioni2018learning}), which show how descriptors based on higher order geometrical attributes turn out more effective on \textit{Random Views} due to the lack of clutter and real sensor noise. As for \textit{KPL-CGF32}, although it performs still overall better than the other descriptors based on point densities, we observe quite a remarkable performance drop compared to the results on the \textit{UWA} dataset, much larger, indeed, than that observed for \textit{KPL-USC}, which shares with \textit{KPL-CGF32} a very similar input representation. This suggests that the non-linear mapping learned by \textit{KPL-CGF32} is highly optimized to tell apart the features belonging to the objects present in the training dataset (\ie{} \textit{UWA}) but turns out quite less effective when applied to unseen features, like those found on the objects belonging to \textit{Random Views}. This \textit{domain shift} issue is a peculiar wick trait of learned features, which may cause them to yield less stable performance across diverse datasets than handcrafted representations.   

\subsection{Surface Registration}
First, it is worth pointing out how, unlike in object recognition settings, in surface registration it is never possible to train any machine learning operator, either detector or descriptor, on the very same objects that would then be processed at test time. Indeed, should one be given either a full 3D model or a set of scans where ground-truth transformations are known, as required to train 3D feature detectors (\ie{} \textit{KPL}) or descriptors (e.g. \textit{CGF-32}), there would be no need to carry out any registration for that object. Surface registration is about stitching together several scans of a new object than one wishes to acquire as a full 3D model. As such, any learning machinery is inherently prone to the domain shift issue.   

\begin{figure}[ht]
	\centering
	\includegraphics[width=0.70\textwidth]{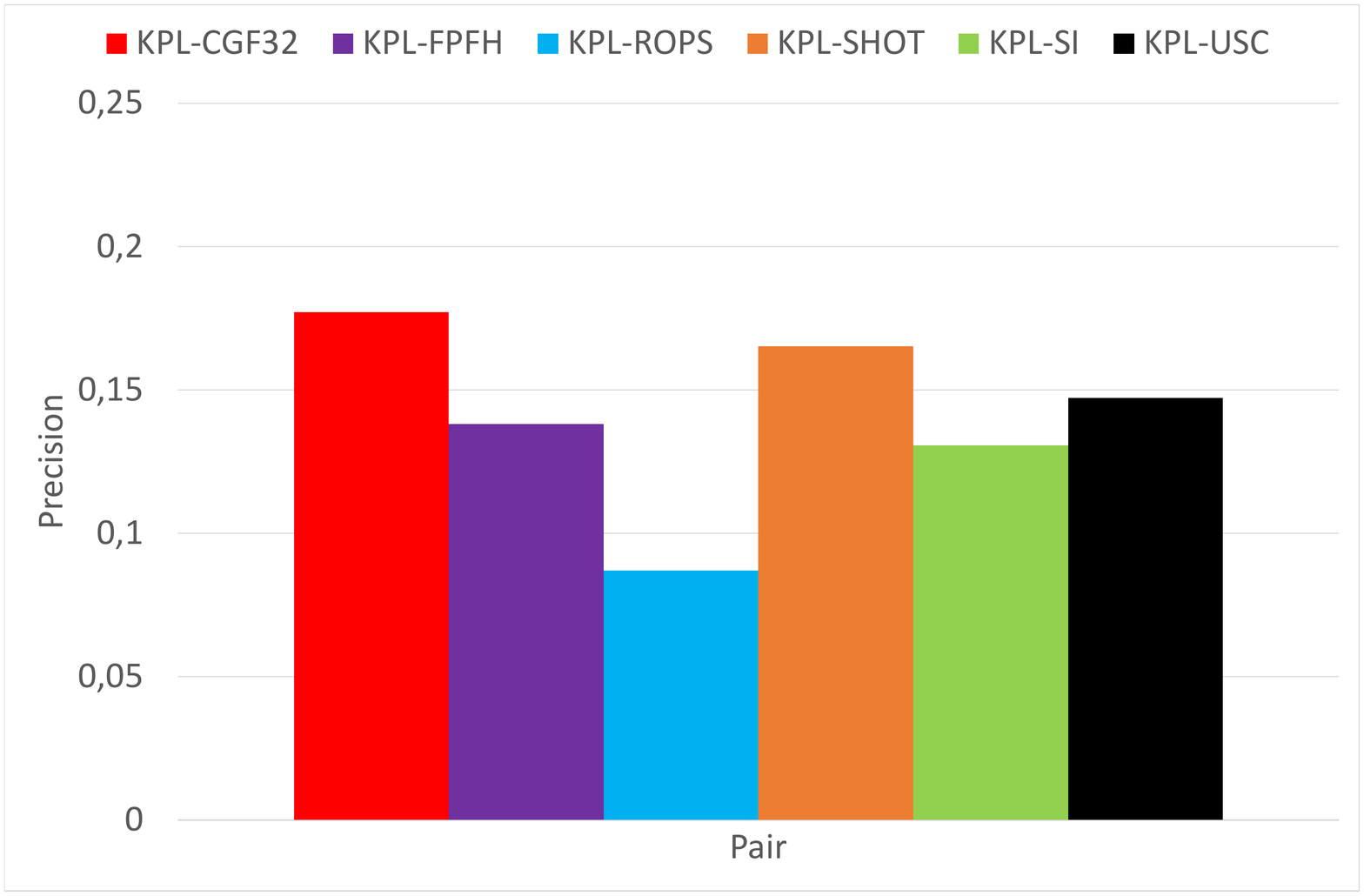}
	\caption{Surface registration results on the \textit{Laser Scan} dataset.}
	\label{fig:SurfRegLaserScan}
\end{figure}

As mentioned in \autoref{SurfaceRegistrationMethodology}, our experiments rely on the \textit{Laser Scan} dataset \cite{khoury2017learning} and follow the split into train and test objects proposed by the authors. As shown in \autoref{fig:SurfRegLaserScan}, when averaging across all test objects, the detector-descriptor pair based on the learned descriptor \textit{CGF-32} provides the best performance. This validates the findings reported in \cite{khoury2017learning}, where the authors introduce \textit{CGF-32} and prove its good registration performance on \textit{Laser Scan}, also in our experimental setting where an optimal detector is learned for every descriptor.

\section{Conclusion and Future Work}

Object recognition settings turn out quite amenable to deploy learned 3D features. Indeed, one can train upon a set of 3D objects available beforehand, \eg due to scanning by some sensor or as CAD models, and then seek to recognize them into scenes featuring occlusions and clutter. These settings allow for learning an highly specialized descriptor alongside its optimal paired detector so to achieve excellent performance. In particular, the learned pair referred to in this paper as \textit{KPL-CGF32} sets the new state of the art in descriptor matching on the \textit{UWA} benchmark dataset. Although the learned representation may not exhibit comparable performance when transferred to unseen objects, in object recognition it is always possible to retrain on the objects at hand to improve performance. An open question left to future work concerns whether the input parametrization deployed by \textit{CGF-32} may enable to learn an highly effective non-linear mapping also in datasets characterized by different nuisances (\eg  \textit{Laser Scan}) or one should better try to learn 3D representations directly from raw data, as vouched by the success of deep learning from image recognition. 
Features based on learned representations, such as \textit{KPL-CGF32}, are quite effective also in surface registration, although  this scenario is inherently more prone to the domain shift issue and, indeed, features based on handcrafted descriptors, like in particular \textit{KPL-SHOT} and \textit{KPL-USC}, turn out very competitive. 

We believe that these findings may pave the way for further research on the recent field  of learned 3D representations, in particular in order to foster addressing domain adaptation issues, a topic investigated more and more intensively in nowadays deep learning literature concerned with image recognition.
Indeed, 3D data are remarkably diverse in nature due to the variety of sensing principles and related technologies and we wittness a lack of large training datasets, \eg at a scale somehow comparable to ImageNet, that may allow learning representations from a rich and varied corpus of 3D models. Therefore, how to effectively transfer learned representations to new scenarios seems a key issue to the success of machine/deep learning in the most challenging 3D Computer Vision tasks.

Finally, \textit{KPL} has established a new framework whereby one can learn an optimal detector for any given descriptor. In this paper we have shown how applying \textit{KPL} to a learned representation (\textit{CGF-32}) leads to particularly effective features (\textit{KPL-CGF32}), in particular when pursuing object recognition. Yet, according to the \textit{KPL} methodology, the descriptor (\eg \textit{CGF-32}) has to be learned before its paired detector: one might be willing to investigate on whether and how a single end-to-end paradigm may allow learning both component jointly so as to further improve performance.


\bibliographystyle{splncs04}
\bibliography{egbib}

\end{document}